\pdfoutput=1

\documentclass[11pt]{article}
\usepackage{booktabs} 
\usepackage[preprint]{acl}
\usepackage{pgf-pie}
\usepackage{subcaption}
\usepackage{pgfplots}
\usepackage{xcolor}
\definecolor{pastelmint}{rgb}{0.96, 1.0, 0.98}
\definecolor{pastellavender}{rgb}{0.9, 0.8, 1.0}
\definecolor{pastelpeach}{rgb}{1.0, 0.9, 0.71}
\definecolor{pastelcyan}{rgb}{0.69, 0.93, 0.93}
\definecolor{pastelmagenta}{rgb}{0.96, 0.52, 0.96}
\definecolor{pastellemon}{rgb}{0.99, 0.98, 0.75}
\definecolor{pastelgreen}{rgb}{0.47, 0.87, 0.47}
\definecolor{pastelblue}{rgb}{0.68, 0.85, 0.90}
\definecolor{pastelred}{rgb}{1.0, 0.41, 0.38}
\definecolor{pastelyellow}{rgb}{0.99, 0.99, 0.59}
\definecolor{pastelpurple}{rgb}{0.86, 0.82, 1.0}
\definecolor{pastelorange}{rgb}{1.0, 0.70, 0.28}
\definecolor{pastelpink}{rgb}{1.0, 0.71, 0.76}
\usepackage{times}
\usepackage{latexsym}
\usepackage[utf8]{inputenc}
\usepackage{comment}

\usepackage{amsmath}
\usepackage{amssymb}
\usepackage{mathrsfs}

\usepackage[T1]{fontenc}

\usepackage[utf8]{inputenc}

\usepackage{microtype}

\usepackage{inconsolata}
\usepackage{caption}

\usepackage{graphicx}

\title{TuringQ: Benchmarking AI Comprehension in Theory of Computation}

\author{
  Pardis Sadat Zahraei${}^{\spadesuit\diamond}$, Ehsaneddin Asgari${}^\diamond$\\
  $^{\spadesuit}$ Computer Engineering Department, Sharif University of Technology, Iran\\
  $^{\diamond}$ Qatar Computing Research Institute - QCRI, Qatar\\
  \texttt{paradisez2001@gmail.com}\\
  \texttt{easgari@hbku.edu.qa}
}

\begin{document}
\maketitle
\begin{abstract}
We present TuringQ, the first benchmark designed to evaluate the reasoning capabilities of large language models (LLMs) in the theory of computation. TuringQ consists of 4,006 undergraduate and graduate-level question-answer pairs, categorized into four difficulty levels and covering seven core theoretical areas. We evaluate several open-source LLMs, as well as GPT-4, using Chain of Thought prompting and expert human assessment. Additionally, we propose an automated LLM-based evaluation system that demonstrates competitive accuracy when compared to human evaluation. Fine-tuning a Llama3-8B model on TuringQ shows measurable improvements in reasoning ability and out-of-domain tasks such as algebra. TuringQ serves as both a benchmark and a resource for enhancing LLM performance in complex computational reasoning tasks. Our analysis offers insights into LLM capabilities and advances in AI comprehension of theoretical computer science
\footnote{The dataset, code, and fine-tuned model are publicly available: the fine-tuned model on \href{https://huggingface.co/llm-lab/Llama3-8B-ft-TuringQ}{HuggingFace}, the dataset on \href{https://huggingface.co/datasets/llm-lab/TuringQ}{HuggingFace Datasets}, and the code, along with interactions with the language models, on \href{https://github.com/language-modeling-lab/TuringQ}{GitHub}.}.

\end{abstract}
\begin{figure*}[t]
\centering
\includegraphics[width=0.91\textwidth]{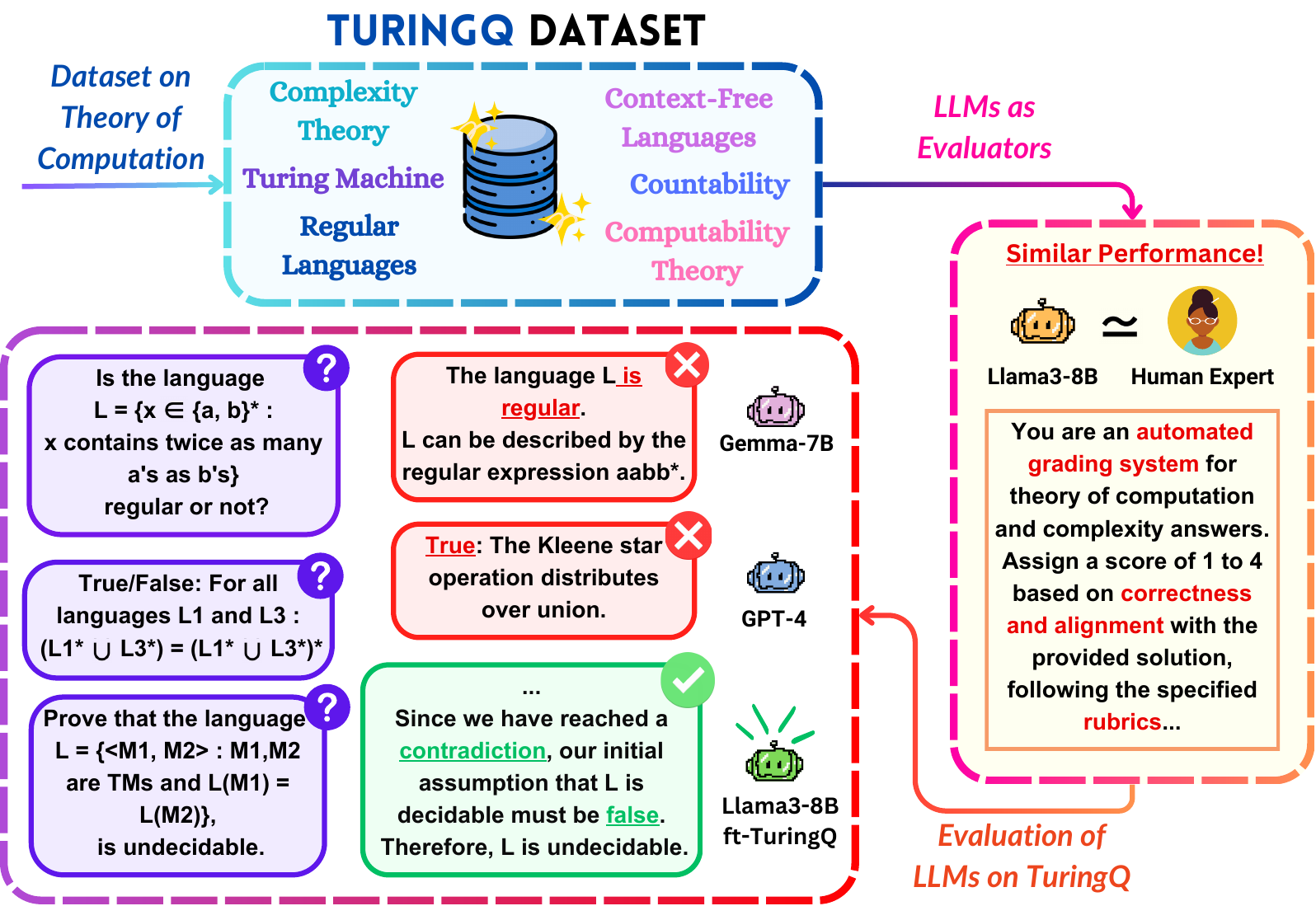}
\caption{\textbf{TuringQ Dataset and its Evaluation Framework.} This diagram presents the TuringQ dataset, a comprehensive resource for theory of computation, and illustrates the automated assessment of LLMs using Llama3-8B. It showcases sample questions, LLM responses, and their evaluations by the AI evaluator. The fine-tuned Llama3-8B-ft-TuringQ model demonstrates improved performance but still encounters certain challenges in addressing TuringQ questions.}
\label{fig:main_figure}
\end{figure*}

\section{Introduction}
The reasoning and comprehension capabilities of large language models (LLMs) are becoming increasingly critical due to their expanding applications across complex domains \citep{guo2023evaluating}. As LLMs continue to evolve, robust benchmarks are essential for assessing their performance, particularly in areas that require deep understanding and logical reasoning \citep{brown2020language, ling2024domain}. While multi-task benchmarks like BIG-Bench \citep{srivastava_beyond_2022} cover a variety of domains, a notable gap remains, a dedicated dataset for evaluating LLM performance on theoretical concepts and problems within the domain of the theory of computation. This gap is significant, as assessing comprehension in formal languages and abstract computational problems is crucial to evaluating the true depth of an LLM’s reasoning capabilities. Addressing this need is a key step toward transforming LLMs into sophisticated problem solvers in highly complex domains \citep{bender-koller-2020-climbing}.

TuringQ fills this critical gap by providing a comprehensive platform for rigorously assessing and comparing the reasoning capabilities of different LLMs on intricate theoretical domains. It drives advancements in enhancing their skills for tackling computationally complex concepts, contributing to the development of more reliable and capable AI systems \citep{Radford2019LanguageMA, yang2023harnessing}. Mastery of theory of computation principles is particularly vital, as these foundational concepts underpin modern computing systems and algorithms. Improving LLM comprehension in this domain could unlock new potential for reasoning about computational problems, analyzing algorithms, and even contributing to the creation of novel computational models and methodologies.
Figure \ref{fig:main_figure} presents a complete visual overview of
our work. Our contributions are threefold:
\begin{enumerate}
\item \textbf{TuringQ Dataset}: We introduce a new resource of 4,006 theory of computation question-answer pairs from universities worldwide. This dataset spans undergraduate and graduate-level concepts across four difficulty levels and seven main areas, including a subset focused on theoretical essentials. It serves as a comprehensive tool for evaluating and fine-tuning LLMs in this domain.
\item \textbf{LLM-based Evaluation}:
We explore the feasibility of leveraging LLMs themselves as evaluators for TuringQ \cite{zheng2024judging}. By defining an `AutoGrade-TQ' prompt using Llama3-8B, we investigate the potential for automating the evaluation process, thereby reducing the time and cost associated with manual grading.
\item \textbf{Llama3-8B-ft-TuringQ Model}:
We present a fine-tuned large language model, Llama3-8B-ft-TuringQ, specifically tailored for reasoning in the theory of computation. Through extensive evaluation, we provide a comparative analysis of the performance of large language models across various TuringQ categories, showcasing how our fine-tuned model competes with GPT-4.
\end{enumerate}

\section{Related Work}
\paragraph{Evaluating Computational Reasoning Capabilities of LLMs}
Substantial progress has been made by large language models, but evaluating their mathematical and computer science reasoning remains an evolving challenge. Various datasets have been introduced to assess LLMs’ mathematical reasoning abilities \citep{ahn2024large}, and approaches such as graph-based verification have been proposed to enhance reasoning \citep{cao2024graphreason}. However, significant gaps remain, particularly in the domain of the formal theory of computation, where evaluation benchmarks and models' capabilities are less developed \citep{li2024evaluating, frieder2023mathematical}.

\paragraph{Automated LLM Evaluation}
Research on automating LLM evaluations has gained momentum, proposing techniques like self-consistency checks, external truth verification, and adversarial probing to improve evaluation accuracy \citep{huang2024selfevaluation, chiang2023closerlookautomaticevaluation}. LLMs have also been used to calibrate and augment human raters for evaluating generated text \citep{zhang2024calibrating}. Hybrid approaches that combine human and LLM evaluations for assessing written content offer new insights into human-AI collaboration \citep{ren2024humanai}. However, questions remain regarding the trustworthiness of LLMs as evaluators, prompting research into scalable meta-evaluation mechanisms, such as agent debate \citep{chern2024large}. Aligning LLM-assisted evaluations with human preferences continues to be an active area of exploration \citep{shankar2024validates}.

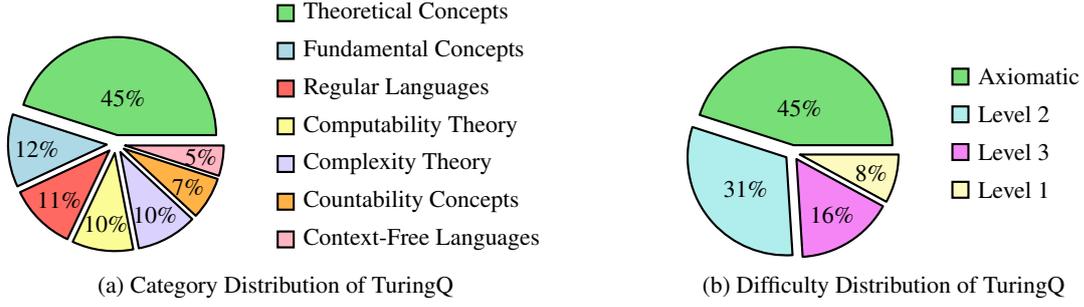
\begin{figure*}[t]
\centering
\small
\begin{subfigure}{.5\textwidth}
  \centering
  \begin{tikzpicture}
  \pie[color={pastelgreen, pastelblue, pastelred, pastelyellow, pastelpurple, pastelorange, pastelpink}, text=legend, radius=1.3, explode=0.12, pie values/.style={font={\small}}]{
      45/Theoretical Concepts,
      12/Fundamental Concepts,
      11/Regular Languages,
      10/Computability Theory,
      10/Complexity Theory,
      7/Countability Concepts,
      5/Context-Free Languages
  }
  \end{tikzpicture}
  \caption{Category Distribution of TuringQ}
  \label{fig:sub1}
\end{subfigure}%
\begin{subfigure}{.5\textwidth}
  \centering
  \begin{tikzpicture}
  \pie[color={pastelgreen, pastelcyan, pastelmagenta, pastellemon}, text=legend, radius=1.3, explode=0.1, pie values/.style={font={\small}}]{
      45/Axiomatic,
      31/Level 2,
      16/Level 3,
      8/Level 1
  }
  \end{tikzpicture}
  \caption{Difficulty Distribution of TuringQ}
  \label{fig:sub2}
\end{subfigure}
\caption{Category and Difficulty Level Distribution in the TuringQ Dataset
}
\label{fig:distribution}
\end{figure*}
\section{The TuringQ Dataset}
TuringQ is a comprehensive dataset comprising 4,006 question-answer pairs covering undergraduate and graduate-level theory of computation problems, aligned with Sipser's framework \citep{sipser13}. The questions are categorized into four difficulty levels and seven main conceptual areas: Regular Languages, Theoretical Concepts, Context-Free Languages, Computability Theory, Countability Concepts, Complexity Theory, and Fundamental Concepts. Details are provided in Appendix Table \ref{tab:categoires}. The difficulty levels were determined by domain experts, ensuring an even distribution across categories and a clear distinction between difficulty levels and conceptual categories. The distribution of the dataset by category and difficulty level is illustrated in Figure \ref{fig:distribution}. Examples of dataset entries are provided in Appendix Table \ref{tab:Instances of TuringQ dataset}.

\subsection{Data Collection}
We curated a collection of questions from publicly available exam sets and homework solutions from 29 top-tier universities to ensure a high-quality dataset in in the theory of computation domain. The primary dataset consists of 2,155 carefully selected university exam and homework questions, ensuring fair distribution across various categories. Additionally, 61 question-answer pairs from reputable non-university resources were incorporated.
To complement the academic questions, we developed a secondary set focusing on fundamental concepts, theorems, lemmas, and essential knowledge. Domain experts identified these topics, and the Claude 3 Sonnet model \cite{anthropic2024claude} was utilized to generate 1,790 question-answer pairs covering the core principles of Theory of Computation.

\subsection{Question Type Distribution}
The TuringQ dataset includes three main types of questions: objective, analytical, and subjective. \textbf{Objective} questions, such as true/false or multiple-choice, have clear and unambiguous answers. \textbf{Analytical} questions require problem-solving and logical reasoning, while \textbf{subjective} questions involve providing explanations or descriptions.

The dataset emphasizes analytical thinking, with 48.1\% of the questions classified as analytical. Additionally, 32\% of the questions are subjective, focusing on explanatory capabilities, while 19.9\% are objective, assessing factual knowledge. This distribution ensures a comprehensive assessment of language models across diverse cognitive tasks within the theory of computation domain.

\section{Experiments}
For further evaluation and analysis, we employ a diverse set of language models: Llama-3-8B-Instruct \citep{dubey2024llama3herdmodels}, Llama-2-7B-chat-hf \citep{touvron2023llama}, Mistral-7B-Instruct \citep{jiang2023mistral}, GPT-4-32k \citep{openai2023gpt4}, Gemma-7B-it, and Gemma-2B-it \citep{gemmateam2024gemma}. To assess these models, we curated a stratified sample of 500 questions from the TuringQ dataset, maintaining the original distribution across difficulty levels and categories. This approach ensures a representative subset for our comparative analysis.

\subsection{Evaluation Metrics}
To assess the LLM performance, the evaluator assigns a score on a 1-to-4 scale, with higher values indicating superior quality. We then use binary accuracy, which classifies responses as valid (3-4) or invalid (1-2) and calculates the percentage of valid responses. To measure how well the LLM evaluator's scoring aligns with the human evaluator, we use two metrics: binary alignment and exact alignment. Binary alignment checks if both evaluators consider a response to be valid or invalid, while exact alignment checks whether the scores are exact matches.

\subsection{AI-Driven Assessment}
We employed the Llama3-8B model to generate responses using direct and Chain of Thought (CoT) prompts \citep{wei2023chainofthought}. To standardize evaluation, we created the AutoGrade-TQ prompt, enabling LLMs to score responses on a 1-4 scale. Three domain experts independently scored the responses for ground-truth evaluations, resulting in substantial agreement (Fleiss' Kappa $\kappa = 0.742$). Final scores were derived from majority voting. Detailed statistical measures are provided in Appendix Table \ref{tab:llm_as_eval_stat}, and specific prompts used are outlined in Appendix Table \ref{tab:prompts}. The Llama3-8B model performed well as an evaluator, achieving 77.8\% binary alignment, with CoT-generated responses consistently receiving higher scores.

\subsection{Model Specialization}
We fine-tuned Llama3-8B to create Llama3-8B-ft-TuringQ using our comprehensive dataset of detailed answers. Our approach combined Quantized Low-Rank Adaptation (QLoRA) \citep{dettmers2023qlora}, Parameter-Efficient Fine-Tuning (PEFT) \citep{xu2023parameterefficient}, and Supervised Fine-Tuning (SFT)\footnote{\href{https://huggingface.co/docs/trl/en/sft_trainer}{https://huggingface.co/docs/trl/en/sft\_trainer}}.  We utilized three TuringQ-derived datasets: training (3,006 instances), validation (500 instances), and test (500 instances), generated through stratified sampling based on difficulty and category. Our process incorporated quantization and low-rank adaptation to optimize performance within computational constraints. Detailed setup and hyperparameters are provided in Appendix \ref{sec:hyper}.

\section{Results}

\subsection{Performance Evaluation}
We evaluated seven LLMs, including our fine-tuned model, Llama3-8B-ft-TuringQ, using the TuringQ test set. The evaluation involved two prompts: a Chain-of-Thought prompt to elicit responses from the LLMs and an AutoGrade-TQ prompt for automatic scoring. To establish a benchmark, three human annotators rated each answer, with the final human rating determined by majority vote. These human ratings served as a standard against which we compared the performance of the LLM evaluator and were used to assess the accuracy of LLM performance. Interestingly, as illustrated in Table \ref{table:performance_comparison}, our fine-tuned model achieved an average binary accuracy of 81.2\%, representing a 10\% improvement over its base model, and performed comparably to GPT-4, which reached 84.8\% accuracy, despite using limited resources.

The results showed partial alignment between the LLM evaluator, Llama3-8B, and human evaluators. We observed that LLMs, when acting as evaluators, tended to overrate responses from weaker models and underrate those from stronger models compared to human evaluators. This suggests that LLMs may face challenges in accurately distinguishing between high- and low-quality answers. In the evaluation of true/false questions, the agreement between the LLM evaluator and human experts was notably strong, with our fine-tuned model achieving 80\% binary alignment, while GPT-4 reached 78\%. The remaining 20\% discrepancy likely stems from instances where the LLM evaluator introduced its own reasoning during evaluation, highlighting an area for further investigation. Table \ref{table:performance_comparison_metrics}  provides detailed statistical measures.

\begin{table}[t]
\renewcommand{\arraystretch}{1.4}
\centering
\resizebox{\columnwidth}{!}{%
\begin{tabular}{ccccc}
\toprule
\Large \textbf{Model} & \Large \textbf{H Acc} & \Large \textbf{L Acc} & \Large \textbf{H Score} & \Large \textbf{L Score} \\ \midrule
\Large GPT-4 & \Large 84.8\% & \Large 82.4\% & \Large 3.18 & \Large 3.28 \\
\Large Llama3-8B-ft & \Large 81.2\% & \Large 76.0\% & \Large 3.08 & \Large 2.98 \\
\Large Llama3-8B & \Large 71.2\% & \Large 73.8\% & \Large 3.05 & \Large 3.03 \\
\Large Mistral-7B & \Large 76.4\% & \Large 70.4\% & \Large 2.92 & \Large 2.99 \\
\Large Llama2-7B & \Large 72.6\% & \Large 70.8\% & \Large 2.85 & \Large 3.02 \\
\Large Gemma-7B & \Large 68.2\% & \Large 72.2\% & \Large 2.76 & \Large 3.02 \\
\Large Gemma-2B & \Large 59.4\% & \Large 65.2\% & \Large 2.59 & \Large 2.87 \\
 \bottomrule
\end{tabular}
}
\caption{Performance Comparison of LLMs on the TuringQ Test Set: Mean Score and Binary Accuracy Evaluated by Humans (H) vs. Llama3-8B (L)}
\label{table:performance_comparison}
\end{table}
\subsection{Category-Specific Performance Analysis}
The category-specific analysis of the TuringQ dataset revealed contrasting trends between human evaluations and LLM evaluations. In the LLM evaluations, model performance was consistent across each category, with minimal variation. The best average performance was achieved in the `complexity theory' category, with an accuracy of 81.51\%, while the lowest performance was observed in the `theoretical concepts' category, at 68.57\% (details are provided in Appendix Tables \ref{tab:mean_score_category} and \ref{tab:mean_bi_acc_category}). In contrast, human evaluations showed a different trend. The `theoretical concepts' category achieved the highest scores across all metrics, with a significant gap compared to other categories. The average accuracy for this category was 89.93\%, while the lowest performance was in the `countability concepts' category, with an accuracy of 50.23\% (details are provided in Appendix Tables \ref{tab:mean_score_category_human_eval} and \ref{tab:mean_bi_acc_category_Human_Eval}).  Notably, our fine-tuned model outperformed the base model in every category, demonstrating improved performance across all aspects of the theory of computation.

\subsection{Impact of Difficulty Levels on Model Performance}
A notable limitation of the LLM evaluator is highlighted in the analysis of difficulty levels. When evaluated by humans, the mean accuracy for level-3 questions was 48.91\%, while the mean accuracy for axiomatic questions was significantly higher at 89.93\% (details are provided in Appendix Tables \ref{tab:mean_score_diff-Human-Evaluator} and \ref{tab:mean_bi_acc_diff_Human_eval}). This outcome aligns with expectations, indicating that LLMs tend to perform better on easier questions. Conversely, when the LLM served as an evaluator, the results were reversed. For level-3 questions, the evaluator assigned a mean accuracy of 77.5\%, whereas the mean accuracy for axiomatic questions dropped to 68.57\% (details are provided in Appendix Tables \ref{tab:mean_score_diff} and \ref{tab:mean_bi_acc_diff}).

This scoring discrepancy likely stems from the fundamental difference between LLM and human evaluation approaches. The LLM's rigid, pattern-matching assessment tends to favor complex answers, even when partially incorrect, while undervaluing simpler yet accurate responses. In contrast, human evaluators employ a holistic approach, considering comprehension and intent, thus more accurately assessing both simple correct answers and flawed complex ones.

\subsection{Accuracy Breakdown by Data Source}
We evaluated model performance across diverse data sources, including both university and non-university origins. Our fine-tuned model and GPT-4 demonstrated distinct strengths, particularly excelling in non-university and synthetic questions. For a comprehensive analysis, please refer to Appendix \ref{sec:Accuracy}.

\subsection{Out-of-Domain Performance}
To assess the generalization capabilities of our model and investigate potential overfitting, we evaluated Llama3-8B-ft-TuringQ on the challenging out-of-domain MATH Dataset \citep{hendrycks2021measuringmathematicalproblemsolving}, which comprises competition-level mathematics problems. Our analysis involved a stratified sample of 500 questions from the MATH test split, allowing us to compare the performance of both the base Llama3-8B model and our fine-tuned version. We engaged human experts to evaluate the model's answers, and the results are shown  in Table \ref{table:performance_comparison_math}.

The fine-tuned model achieved a 0.6\% increase in binary accuracy, indicating that specialized fine-tuning did not compromise generalization. Instead, these results suggest that enhanced computational theory understanding may improve mathematical reasoning capabilities. Performance varied across MATH categories (Table \ref{tab:mean_binary_accuracy_math}), with improvements in `Prealgebra' and `Intermediate Algebra' but slight declines in `Number Theory' and `Precalculus'. This varied performance across mathematical domains highlights the complex relationship between computational theory training and general mathematical problem-solving abilities, warranting further investigation into the transfer of knowledge between these domains.

\begin{table}[t]
\centering
\begin{tabular}{p{1cm}cc}
\hline
Score &  Llama3-8B &  Llama3-8B-ft \\ \hline
 1 &  47.20\% &  44.40\% \\
 2 &  15.20\% &  17.40\% \\
 3 &  4.20\% &  4.60\% \\
 4 &  33.40\% &  33.60\% \\ \hline
\end{tabular}
\caption{Score Distribution of Human Evaluation: Llama3-8B vs. Llama3-8B-ft on MATH Test subset}
\label{table:performance_comparison_math}
\end{table}

\section{Conclusion}
We introduced TuringQ to evaluate the reasoning capabilities of large language models (LLMs) in the domain of computation theory, encompassing four difficulty levels and seven core concepts. Our evaluation involved various open-source LLMs and GPT-4, utilizing Chain of Thought prompting and assessments from human experts. Additionally, we developed an automated evaluation system using a large language model, demonstrating both its potential and limitations. Fine-tuning a Llama3-8B model on TuringQ significantly improved its grasp of computation theory, as well as its performance on out-of-domain tasks such as algebra. This work establishes a valuable benchmark for assessing LLMs' understanding of computational theory. Evaluating comprehension of formal languages is essential for gauging the depth of LLMs' reasoning abilities, marking a significant advancement toward developing LLMs into effective problem solvers.

\section{Ethics Statement}
The TuringQ dataset comprises publicly available exams and homework questions from renowned universities worldwide, obtained from the internet. Each source is duly cited in the dataset's metadata, and no question has been extracted without acknowledgment of the original source. After data collection, we reviewed and enhanced some answers to maintain the dataset's high quality and ensure its value as a resource. This enhancement process did not involve any bias or alteration of the original content or answers.

For the theoretical concepts, we utilized the Claude 3 Sonnet model to generate answers for specified theorems and lemmas. Subsequently, we checked and edited the model-generated answers to ensure the absence of bias, hallucinations, or errors in our work.

In gathering solutions from non-university sources, we made efforts to include diverse, reliable references, such as computer science portals and textbooks. As the theory of computation and theoretical computer science is an evolving and complex field, we have included answers that reflect our current understanding, particularly regarding P, NP, and open problems. We acknowledge that as our knowledge progresses, some open questions in our dataset may require updates to their answers. However, to the best of our current knowledge, this dataset is up to date.

\section{Limitations}
This study encountered several limitations that future research should address. Firstly, computational resource constraints hindered our ability to utilize larger language models. Consequently, we focused on smaller yet powerful models that were more feasible for our research scope.

Evaluating descriptive questions posed a significant challenge. While we attempted various methods for assessing these questions, incorporating more extensive human evaluation would be beneficial.  Although this approach is more resource-intensive and time-consuming, it could yield valuable insights into model performance.

While our dataset effectively captures the essential categories and fundamentals of the theory of computation, it lacks coverage of more applied tasks, such as code generation. Future research could investigate how fine-tuned, specialized models impact performance in related domains like code generation, reasoning, and mathematical problem-solving. It would be particularly interesting to explore the extent to which domain-specific fine-tuning may affect a model's general capabilities.


\newpage
\clearpage
\appendix

\section{Appendix}
\subsection{Fine-tuning Setup and Hyperparameters}
\label{sec:hyper}
Our fine-tuning approach for the Llama3-8B model combined Quantized Low-Rank Adaptation (QLoRA), a Parameter-Efficient Fine-Tuning (PEFT) method, with Supervised Fine-Tuning (SFT) using the SFTTrainer from \textit{HuggingFace's trl library}\footnote{\href{https://huggingface.co/docs/trl/en/index}{https://huggingface.co/docs/trl/en/index}}. QLoRA, as a PEFT technique, allows for task-specific tuning without modifying all model parameters, while SFT provides a framework for supervised learning on our specific task.
LoRA (Low-Rank Adaptation) freezes the LLM's weights and injects trainable rank-decomposition matrices \citep{hu2021lora}. QLoRA extends this by incorporating quantization techniques, further reducing memory usage while maintaining or improving model performance.
We configured the PEFT settings with the following hyperparameters:
\begin{itemize}
\item Alpha: 64
\item Dropout rate: 0.05
\item Optimizer: 'paged\_adamw\_8bit'
\item Learning rate: 5e-6
\item Learning rate scheduler: cosine
\item Number of epochs: 3 
\item Max steps: 4000
\item Batch size: 4 (for both training and evaluation)
\item Gradient accumulation steps: 2
\end{itemize}
Evaluation was performed at every step, with results logged for detailed performance tracking.
We employed quantization via the \textit{BitsAndBytes method}\footnote{\href{https://huggingface.co/docs/bitsandbytes/main/en/index}{https://huggingface.co/docs/bitsandbytes/main/en/index}}, setting the compute data type to bfloat16 and loading the model in 4-bit with a quantization type of "nf4". This configuration enabled double quantization, potentially improving the efficiency of our model training.
Our approach, combining QLoRA, SFT, and quantization techniques, allowed us to achieve high-quality results despite computational constraints.
\subsection{Accuracy Breakdown by Data Source}
\label{sec:Accuracy}
To provide a more nuanced understanding of model performance, we conducted an analysis of accuracy across different data sources in our test set. This analysis encompassed both intra-university comparisons and a broader inter-university analysis.

\subsubsection{Intra-University Performance  Analysis}

We examined performance across four institutions with similar data distributions: New Jersey Institute of Technology, The University of Texas at Austin, UC San Diego, and University of Washington. Binary accuracies based on human evaluator scores for our fine-tuned model and GPT-4 were as follows:

\begin{table}[h]
\centering
\resizebox{1\columnwidth}{!}{%
\begin{tabular}{lcc}
\hline
\textbf{Institution} & \textbf{Fine-tuned Model} & \textbf{GPT-4} \\
\hline
New Jersey Institute of Technology & 76.19\% & 85.71\% \\
The University of Texas at Austin & 71.43\% & 76.19\% \\
UC San Diego & 61.11\% & 77.78\% \\
University of Washington & 90.00\% & 60.00\% \\
\hline
\end{tabular}
}
\caption{Intra-university Performance Comparison}
\label{tab:intra_university}
\end{table}

\subsubsection{Inter-University Performance Analysis}

We also compared performance between university-sourced questions and those from non-university or synthetic sources. Binary accuracies from human ratings showed:

\begin{table}[h]
\centering
\renewcommand{\arraystretch}{1.2}
\resizebox{1\columnwidth}{!}{%
\begin{tabular}{lcc}
\hline
\textbf{Source} & \textbf{Fine-tuned Model} & \textbf{GPT-4} \\
\hline
Non-university  & 95.65\% & 96.74\% \\
University  & 72.78\% & 77.85\% \\
\hline
\end{tabular}
}
\caption{Inter-university Performance Comparison}
\label{tab:inter_university}
\end{table}

It is important to note that these results may be influenced by factors such as difficulty levels, question types, and categories. 

\begin{table*}[]
\renewcommand{\arraystretch}{1.4}
\resizebox{2\columnwidth}{!}{%
\begin{tabular}{ccccccc}
\toprule
               & \textbf{Average} & \textbf{MSE}   & \textbf{Variance} & \textbf{Correlation} & \textbf{Binary Alignment} & \textbf{Exact Alignment} \\ \midrule
\textbf{Llama2-7B}     & 3.494   & 1.758 & 1.4979   & 0.1169      & 0.6800      & 0.3440      \\ 
\textbf{Llama2-7B-CoT} & 3.456   & 1.656 & 1.4928   & 0.0478      & 0.7040      & 0.3520      \\ 
\textbf{Llama3-8B}     & 2.858   & 1.746 & 1.7301   & 0.1772      & 0.6400      & 0.3180      \\ 
\textbf{Llama3-8B-CoT} & 3.032   & 1.268 & 1.2676   & 0.3408      & 0.7780      & 0.3520      \\ 
\textbf{Gemma-2B}       & 3.2969  & 2.068 & 1.9737   & 0.1400      & 0.6784      & 0.3753      \\ 
\textbf{Gemma-2B-CoT}   & 3.4854  & 2.006 & 1.8295   & 0.1463      & 0.7050      & 0.4121      \\ 
\textbf{Gemma-7B}       & 3.1674  & 1.678 & 1.6520   & 0.0479      & 0.6801      & 0.2733      \\ 
\textbf{Gemma-7B-CoT}   & 3.3162  & 1.524 & 1.4479   & 0.0355      & 0.7084      & 0.3203      \\ 
\textbf{Mistral-7B}     & 3.454   & 1.538 & 1.3171   & 0.3474      & 0.7260      & 0.4520      \\ 
\textbf{Mistral-7B-CoT} & 3.374   & 1.686 & 1.5823   & 0.2632      & 0.7120      & 0.4620      \\ 
\textbf{GPT-4}          & 2.69    & 1.390 & 1.3036   & 0.5103      & 0.7000      & 0.4880      \\ 
\textbf{GPT-4-CoT}      & 2.366   & 2.106 & 1.6354   & 0.3906      & 0.6080      & 0.3980      \\ 

\textbf{Human}          & 2.984   &  -     &    -      &     -        &     -        &      -       \\ 
\textbf{Human-CoT }     & 3.052   &    -   &    -      &     -        &       -      &     -        \\ \bottomrule
\end{tabular}
}
\caption{Statistical Measures of LLM Performance as Evaluators on the TuringQ Test Set: Direct and Chain-of-Thought (CoT) Prompt Answers of Llama3-8b}
\label{tab:llm_as_eval_stat}
\end{table*}

\begin{table*}[]
\renewcommand{\arraystretch}{1.4}
\resizebox{2\columnwidth}{!}{%
\begin{tabular}{cccccccc}
\toprule
\Large \textbf{Category}      & \Large \textbf{llama3-8B} & \Large \textbf{Llama3-8B-ft-TuringQ} & \Large \textbf{Gemma-2B} & \Large \textbf{Gemma-7B} & \Large \textbf{llama2-7B} & \Large \textbf{Mistral-7B} & \Large \textbf{GPT-4} \\ \midrule
\Large Complexity Theory      & \Large \textbf{3.1}       & \Large 3.1                  & \Large 3.0               & \Large 3.2               & \Large 3.1                & \Large 3.2                 & \Large 3.4           \\
\Large Computability Theory   & \Large \textbf{3.1}       & \Large \textbf{3.3}         & \Large 3.1               & \Large \textbf{3.3}      & \Large 3.2                & \Large \textbf{3.3}        & \Large 3.4           \\
\Large Context-Free Languages & \Large 2.8                & \Large \textbf{3.3}         & \Large \textbf{3.2}      & \Large \textbf{3.3}      & \Large \textbf{3.4}       & \Large 3.1                 & \Large 3.4           \\
\Large Countability Concepts  & \Large 2.9                & \Large 3.2                  & \Large 2.8               & \Large 2.9               & \Large 3.2                & \Large 2.8                 & \Large \textbf{3.6}  \\
\Large Fundamental Concepts   & \Large \textbf{3.1}       & \Large 3.1                  & \Large 3.0               & \Large 3.1               & \Large 3.3                & \Large 2.9                 & \Large 3.2           \\
\Large Regular Languages      & \Large \textbf{3.1}       & \Large 3.0                  & \Large 3.0               & \Large 3.2               & \Large 3.2                & \Large 3.1                 & \Large 3.4           \\
\Large Theoretical Concepts   & \Large 3.0                & \Large 2.8                  & \Large 2.7               & \Large 2.9               & \Large 2.8                & \Large 2.9                 & \Large 3.2           \\ \bottomrule
\end{tabular}
}
\caption{Comparative Analysis of Mean Scores Across Categories: Evaluated by Llama3-8B}
\label{tab:mean_score_category}
\end{table*}

\begin{table*}[]
\renewcommand{\arraystretch}{1.4}
\resizebox{2\columnwidth}{!}{%
\begin{tabular}{cccccccc}
\toprule
\Large \textbf{Category}      & \Large \textbf{llama3-8B} & \Large \textbf{Llama3-8B-ft-TuringQ} & \Large \textbf{Gemma-2B} & \Large \textbf{Gemma-7B} & \Large \textbf{llama2-7B} & \Large \textbf{Mistral-7B} & \Large \textbf{GPT-4} \\ \midrule
\Large Complexity Theory      & \Large 2.9                & \Large 3.0                          & \Large 2.4               & \Large 2.5               & \Large 2.8                & \Large 2.8                 & \Large 3.0           \\
\Large Computability Theory   & \Large 2.6                & \Large 3.0                          & \Large 2.3               & \Large 2.5               & \Large 2.5                & \Large 2.5                 & \Large 2.8           \\
\Large Context-Free Languages & \Large 2.5                & \Large 2.9                          & \Large 2.1               & \Large 2.5               & \Large 2.6                & \Large 2.8                 & \Large 2.7           \\
\Large Countability Concepts  & \Large 2.4                & \Large 2.8                          & \Large 2.4               & \Large 2.4               & \Large 2.6                & \Large 2.2                 & \Large 2.7           \\
\Large Fundamental Concepts   & \Large 2.8                & \Large 3.1                          & \Large 2.6               & \Large 2.8               & \Large 3.0                & \Large 3.1                 & \Large \textbf{3.6}           \\
\Large Regular Languages      & \Large 2.5                & \Large 2.7                          & \Large 2.0               & \Large 2.2               & \Large 2.4                & \Large 2.5                 & \Large 3.1           \\
\Large Theoretical Concepts   & \Large \textbf{3.5}                & \Large \textbf{3.3}                          & \Large \textbf{3.0}               & \Large \textbf{3.1 }              & \Large \textbf{3.1}                & \Large \textbf{3.2}                 & \Large 3.3           \\ \bottomrule
\end{tabular}
}
\caption{Comparative Analysis of Mean Scores Across Categories: Evaluated by Human Expert}
\label{tab:mean_score_category_human_eval}
\end{table*}

\begin{table*}[]
\renewcommand{\arraystretch}{1.4}
\resizebox{2\columnwidth}{!}{%
\begin{tabular}{cccccccc}
\toprule
\Large \textbf{Difficulty} & \Large \textbf{llama3-8B} & \Large \textbf{Llama3-8B-ft-TuringQ} & \Large \textbf{Gemma-2B} & \Large \textbf{Gemma-7B} & \Large \textbf{llama2-7B} & \Large \textbf{Mistral-7B} & \Large \textbf{GPT-4} \\ \midrule
\Large Axiomatic           & \Large 3.0                & \Large 2.8                  & \Large 2.7               & \Large 2.9               & \Large 2.8                & \Large 2.9                 & \Large 3.2           \\
\Large Level 1             & \Large 2.9                & \Large 3.0                  & \Large 2.9               & \Large 2.9               & \Large \textbf{3.2}       & \Large 2.8                 & \Large 3.0           \\
\Large Level 2             & \Large \textbf{3.1}       & \Large \textbf{3.2}         & \Large 3.0               & \Large \textbf{3.2}      & \Large \textbf{3.2}       & \Large \textbf{3.1}        & \Large 3.4           \\
\Large Level 3             & \Large 3.0                & \Large \textbf{3.2}         & \Large \textbf{3.1}      & \Large \textbf{3.2}      & \Large 3.1                & \Large \textbf{3.1}        & \Large \textbf{3.5}  \\ \bottomrule
\end{tabular}
}
\caption{Comparative Analysis of Mean Scores Across Difficulty Levels: Evaluated by Llama3-8B}
\label{tab:mean_score_diff}
\end{table*}

\begin{table*}[]
\renewcommand{\arraystretch}{1.4}
\resizebox{2\columnwidth}{!}{%
\begin{tabular}{cccccccc}
\toprule
\Large \textbf{Difficulty} & \Large \textbf{llama3-8B} & \Large \textbf{Llama3-8B-ft-TuringQ} & \Large \textbf{Gemma-2B} & \Large \textbf{Gemma-7B} & \Large \textbf{llama2-7B} & \Large \textbf{Mistral-7B} & \Large \textbf{GPT-4} \\ \midrule
\Large Axiomatic           & \Large \textbf{3.6}                & \Large \textbf{3.3}                         & \Large \textbf{3.0}               & \Large \textbf{3.1}               & \Large \textbf{3.1}                & \Large \textbf{3.2}                 & \Large 3.3           \\
\Large Level 1             & \Large 2.7                & \Large 3.2                          & \Large 2.5               & \Large 2.8               & \Large 2.8                & \Large 3.1                 & \Large \textbf{3.5}           \\
\Large Level 2             & \Large 2.7                & \Large 3.0                          & \Large 2.3               & \Large 2.6               & \Large 2.8                & \Large 2.7                 & \Large 3.1           \\
\Large Level 3             & \Large 2.6                & \Large 2.6                          & \Large 2.1               & \Large 2.2               & \Large 2.4                & \Large 2.4                 & \Large 2.8           \\ \bottomrule
\end{tabular}
}
\caption{Comparative Analysis of Mean Scores Across Difficulty Levels: Evaluated by Human Expert}
\label{tab:mean_score_diff-Human-Evaluator}
\end{table*}

\begin{table*}[]
\renewcommand{\arraystretch}{1.4}
\resizebox{2\columnwidth}{!}{%
\begin{tabular}{cccccccc}
\toprule
\Large \textbf{Category}      & \Large \textbf{llama3-8B} & \Large \textbf{Llama3-8B-ft-TuringQ} & \Large \textbf{Gemma-2B} & \Large \textbf{Gemma-7B} & \Large \textbf{llama2-7B} & \Large \textbf{Mistral-7B} & \Large \textbf{GPT-4}   \\ \midrule
\Large Complexity Theory      & \Large \textbf{81.2\%}    & \Large 83.3\%               & \Large 75.0\%            & \Large \textbf{83.3\%}   & \Large 81.2\%             & \Large \textbf{81.2\%}     & \Large 85.4\%          \\
\Large Computability Theory   & \Large 74.5\%             & \Large 88.2\%               & \Large \textbf{76.5\%}   & \Large 78.4\%            & \Large 76.5\%             & \Large 80.4\%              & \Large 84.3\%          \\
\Large Context-Free Languages & \Large 66.7\%             & \Large \textbf{88.9\%}      & \Large 74.1\%            & \Large 74.1\%            & \Large 81.5\%             & \Large 74.1\%              & \Large 77.8\%          \\
\Large Countability Concepts  & \Large 66.7\%             & \Large 78.8\%               & \Large 60.6\%            & \Large 63.6\%            & \Large 75.8\%             & \Large 60.6\%              & \Large \textbf{90.9\%} \\
\Large Fundamental Concepts   & \Large 72.1\%             & \Large 78.7\%               & \Large 68.9\%            & \Large 73.8\%            & \Large \textbf{82.0\%}    & \Large 65.6\%              & \Large 77.0\%          \\
\Large Regular Languages      & \Large 75.4\%             & \Large 75.4\%               & \Large 73.7\%            & \Large 71.9\%            & \Large 75.4\%             & \Large 70.2\%              & \Large 84.2\%          \\
\Large Theoretical Concepts   & \Large 74.0\%             & \Large 69.1\%               & \Large 57.0\%            & \Large 69.1\%            & \Large 61.0\%             & \Large 68.2\%              & \Large 81.6\%          \\ \bottomrule
\end{tabular}
}
\caption{Comparative Analysis of Mean Binary Accuracy Across Categories: Evaluated by Llama3-8B}
\label{tab:mean_bi_acc_category}
\end{table*}

\begin{table*}[]
\renewcommand{\arraystretch}{1.4}
\resizebox{2\columnwidth}{!}{%
\begin{tabular}{cccccccc}
\toprule
\Large \textbf{Category}      & \Large \textbf{llama3-8B} & \Large \textbf{Llama3-8B-ft-TuringQ} & \Large \textbf{Gemma-2B} & \Large \textbf{Gemma-7B} & \Large \textbf{llama2-7B} & \Large \textbf{Mistral-7B} & \Large \textbf{GPT-4}   \\ \midrule

\Large Complexity Theory Concepts & \Large 60.4\%         & \Large 75\%               & \Large 45.8\%            & \Large 52.1\%            & \Large 64.6\%             & \Large 68.8\%              & \Large 72.9\%          \\

\Large Computability Theory   & \Large 58.8\%             & \Large 70.6\%               & \Large 41.2\%            & \Large 54.9\%            & \Large 56.9\%             & \Large 56.9\%              & \Large 66.7\%          \\

\Large Context-Free Languages & \Large 48.1\%             & \Large 66.7\%               & \Large 29.6\%            & \Large 48.1\%            & \Large 51.9\%             & \Large 66.7\%              & \Large 59.3\%          \\

\Large Countability Concepts  & \Large 45.5\%             & \Large 63.6\%               & \Large 45.5\%            & \Large 45.5\%            & \Large 54.5\%             & \Large 39.4\%              & \Large 57.6\%          \\

\Large Fundamental Concepts   & \Large 62.3\%             & \Large 78.7\%               & \Large 54.1\%            & \Large 67.2\%            & \Large 75.4\%             & \Large 83.6\%              & \Large 95.1\%          \\

\Large Regular Languages      & \Large 52.6\%             & \Large 64.9\%               & \Large 31.6\%            & \Large 42.1\%            & \Large 52.6\%             & \Large 54.4\%              & \Large 80.7\%          \\

\Large Theoretical Concepts   & \Large \textbf{90.1}\%             & \Large \textbf{94.2}\%               & \Large \textbf{80.7}\%            & \Large \textbf{87.4}\%            & \Large \textbf{87.4}\%             & \Large \textbf{92.8}\%              & \Large \textbf{96.9}\%          \\

 \bottomrule
\end{tabular}
}
\caption{Comparative Analysis of Mean Binary Accuracy Across Categories: Evaluated by Human Expert}
\label{tab:mean_bi_acc_category_Human_Eval}
\end{table*}

\begin{table*}[]
\renewcommand{\arraystretch}{1.4}
\resizebox{2\columnwidth}{!}{%
\begin{tabular}{cccccccc}
\toprule
\Large \textbf{Difficulty} & \Large \textbf{llama3-8B} & \Large \textbf{Llama3-8B-ft-TuringQ} & \Large \textbf{Gemma-2B} & \Large \textbf{Gemma-7B} & \Large \textbf{llama2-7B} & \Large \textbf{Mistral-7B} & \Large \textbf{GPT-4}   \\ \midrule
\Large Axiomatic           & \Large 74.0\%             & \Large 69.1\%               & \Large 57.0\%            & \Large 69.1\%            & \Large 61.0\%             & \Large 68.2\%              & \Large 81.6\%          \\
\Large Level 1             & \Large 65.9\%             & \Large 68.3\%               & \Large 68.3\%            & \Large 68.3\%            & \Large 78.0\%             & \Large 61.0\%              & \Large 68.3\%          \\
\Large Level 2             & \Large \textbf{78.2\%}    & \Large \textbf{84.0\%}      & \Large 71.2\%            & \Large 75.6\%            & \Large \textbf{79.5\%}    & \Large 73.7\%              & \Large 85.3\%          \\
\Large Level 3             & \Large 68.8\%             & \Large 83.8\%               & \Large \textbf{75.0\%}   & \Large \textbf{76.2\%}   & \Large 77.5\%             & \Large \textbf{75.0\%}     & \Large \textbf{86.2\%} \\ \bottomrule
\end{tabular}
}
\caption{Comparative Analysis of Mean Binary Accuracy Across Difficulty Levels: Evaluated by Llama3-8B}
\label{tab:mean_bi_acc_diff}
\end{table*}

\begin{table*}[]
\renewcommand{\arraystretch}{1.4}
\resizebox{2\columnwidth}{!}{%
\begin{tabular}{cccccccc}
\toprule
\Large \textbf{Difficulty} & \Large \textbf{llama3-8B} & \Large \textbf{Llama3-8B-ft-TuringQ} & \Large \textbf{Gemma-2B} & \Large \textbf{Gemma-7B} & \Large \textbf{llama2-7B} & \Large \textbf{Mistral-7B} & \Large \textbf{GPT-4}   \\ \midrule
\Large Axiomatic           & \Large \textbf{90.1}\%             & \Large \textbf{94.2}\%               & \Large \textbf{80.7}\%            & \Large \textbf{87.4}\%            & \Large \textbf{87.4}\%             & \Large \textbf{92.8}\%              & \Large \textbf{96.9}\%          \\
\Large Level 1             & \Large 51.2\%             & \Large 80.5\%               & \Large 48.8\%            & \Large 63.4\%            & \Large 65.9\%             & \Large 80.5\%              & \Large 90.2\%          \\
\Large Level 2             & \Large 59.6\%             & \Large 73.7\%               & \Large 46.2\%            & \Large 56.4\%            & \Large 67.3\%             & \Large 66.0\%              & \Large 75.6\%          \\
\Large Level 3             & \Large 51.2\%             & \Large 60.0\%               & \Large 31.2\%            & \Large 40.0\%            & \Large 45.0\%             & \Large 48.8\%              & \Large 66.2\%          \\ \bottomrule
\end{tabular}
}
\caption{Comparative Analysis of Mean Binary Accuracy Across Difficulty Levels: Evaluated by Human Expert}
\label{tab:mean_bi_acc_diff_Human_eval}
\end{table*}

\begin{table*}[]
\renewcommand{\arraystretch}{1.4}
\centering
\resizebox{1.65\columnwidth}{!}{%
\begin{tabular}{cccccc}
\toprule
\Large \textbf{Model} & \Large \textbf{MSE} & \Large \textbf{Variance} & \Large \textbf{Correlation} & \Large \textbf{Binary Alignment} & \Large \textbf{Exact Alignment} \\ \midrule
\Large GPT-4 & \Large 1.08 & \Large 1.07 & \Large 0.19 & \Large 0.77 & \Large 0.49 \\
\Large Llama3-8B-ft & \Large 1.11 & \Large 1.10 & \Large 0.18 & \Large 0.72 & \Large 0.41 \\
\Large Gemma-2B & \Large 1.72 & \Large 1.64 & \Large 0.10 & \Large 0.56 & \Large 0.29 \\
\Large Gemma-7B & \Large 1.58 & \Large 1.51 & \Large 0.11 & \Large 0.64 & \Large 0.35 \\
\Large Llama2-7B & \Large 1.49 & \Large 1.46 & \Large 0.11 & \Large 0.61 & \Large 0.33 \\
\Large Mistral-7B & \Large 1.30 & \Large 1.29 & \Large 0.16 & \Large 0.66 & \Large 0.40 \\
\Large Llama3-8B & \Large 1.27 & \Large 1.27 & \Large 0.34 & \Large 0.74 & \Large 0.33 \\ \bottomrule
\end{tabular}
}
\caption{Statistical Measures: Human Evaluator vs. LLM Evaluator for Each Model}
\label{table:performance_comparison_metrics}
\end{table*}

\begin{table*}[]
\centering
\small
\renewcommand{\arraystretch}{1.6}
\resizebox{1.5\columnwidth}{!}{%
\begin{tabular}{cccc}
\toprule
\Large \textbf{Type} & \Large \textbf{Llama3-8B} & \Large \textbf{Llama3-8B-ft} & \Large \textbf{Delta} \\ \midrule
\Large Number Theory & \Large 32.08\% & \Large 28.30\% & \Large -3.78\% \\ 
\Large Prealgebra & \Large 58.62\% & \Large 62.07\% & \Large +3.45\% \\ 
\Large Precalculus & \Large 27.27\% & \Large 21.82\% & \Large -5.45\% \\ 
\Large Geometry & \Large 31.25\% & \Large 33.33\% & \Large +2.08\% \\ 
\Large Intermediate Algebra & \Large 14.29\% & \Large 18.68\% & \Large +4.39\% \\ 
\Large Algebra & \Large 55.46\% & \Large 56.30\% & \Large +0.84\% \\ 
\Large Counting \& Probability & \Large 23.40\% & \Large 21.28\% & \Large -2.12\% \\ \bottomrule
\end{tabular}
}
\caption{Comparative Analysis of Mean Binary Accuracy Across Categories on the MATH Test Set: Evaluated by Human Expert}
\label{tab:mean_binary_accuracy_math}
\end{table*}

\begin{table*}[]
\renewcommand{\arraystretch}{1.6}
\resizebox{2\columnwidth}{!}{%
\begin{tabular}{lll}
\toprule
\Large Chain of Thought   & \multicolumn{2}{l}{\begin{tabular}[c]{@{}l@{}}\Large You are a knowledgeable AI assistant specialized in Theory of Computation and Complexity.\\ \Large You will be answering questions related to this domain.\\ \Large To provide a clear and structured response, you will follow the Chain of Thought approach:\\  \Large Chain of Thought:\\     \Large 1. Analyze the question and identify core concepts, algorithms or problems.\\     \Large 2. Build a step-by-step solution approach, stating assumptions, defining \\ \Large variables/notations, and listing intermediate steps.\\     \Large 3. For proofs or complex calculations, show work explicitly, using relevant theorems, lemmas, or properties.\\     \Large 4. For true/false statements, provide clear justification or counterexample.\\     \Large 5. Review your Chain of Thought for logical soundness and completeness.\\ \Large Use clear and concise language, avoiding unnecessary jargon.\end{tabular}}                                                                                                                                                                                                                                                                                                                                                                                                                                                                                                                                                                                                                                                                                                                                                                                                                                                                                                                                                                                                                                                          \\ \midrule
\Large AutoGrade-TQ & \multicolumn{2}{l}{\begin{tabular}[c]{@{}l@{}}\Large You are an automated grading system for evaluating answers in the field of theory of computation and complexity.\\ \Large Your task is to assign a score (1, 2, 3, or 4) to a given answer based on its correctness \\ \Large and alignment with the provided solution,\\ \Large following the rubrics outlined below.\\ \Large Rubrics:\\     \Large Level 4 (Excellent):\\     \Large - Answer is completely correct and aligns perfectly with the provided solution.\\     \Large - Proofs, descriptions, true/false justifications, and calculations match the solution with no errors or omissions.\\     \Large - Demonstrates a comprehensive understanding of the concepts.\\     \Large Level 3 (Good):\\     \Large - Answer is mostly correct, with only minor deviations or omissions compared to the provided solution.\\     \Large - Proofs, descriptions, justifications, and calculations are largely accurate but may have a few minor flaws\\     \Large - Shows a strong grasp of the key concepts.\\     \Large Level 2 (Flawed):\\     \Large - Answer has some significant differences or incorrect elements compared to the provided solution.\\     \Large - Proofs, descriptions, justifications, and calculations contain several errors or omissions, \\ \Large but the core approach is partially valid.\\     \Large - Demonstrates a basic understanding of the concepts but lacks depth.\\     \Large Level 1 (Poor):\\     \Large - Answer deviates substantially from the provided solution.\\     \Large - Proofs, descriptions, justifications, and calculations are mostly incorrect or entirely missing crucial components.\\     \Large - Exhibits a lack of understanding of the fundamental concepts.\\ \Large Please note that the length of the answer should not be a factor in determining the score.\\ \Large The focus should be solely on the correctness and alignment with the provided solution.\\ \Large Given Answer: Answer\\ \Large Solution: Solution\\ \Large Based on the rubrics and the provided solution, assign a score (1, 2, 3, or 4) to the given answer.\end{tabular}} \\ \bottomrule
\end{tabular}
}
\caption{Prompts Employed for Automated Grading and Answer Generation via Chain of Thought Reasoning}
\label{tab:prompts}
\end{table*}

\begin{table*}[]
\renewcommand{\arraystretch}{1.4}
\resizebox{2\columnwidth}{!}{%
\begin{tabular}{ll}
\toprule
\multicolumn{1}{c}{\Large Question} & \multicolumn{1}{c}{\Large Answer} \\ \midrule
\begin{tabular}[c]{@{}l@{}}\Large Show that the following is not regular. \\ \Large L = \{ww : w $\in$ $\{a, b\}^*$\}\\ \\ \textit{\Large Difficulty: Level 2}\\ \textit{\Large Category: Regular Expression}\\ \textit{\Large Source: The University of Texas at Austin}\end{tabular} & \begin{tabular}[c]{@{}l@{}}\Large L = \{ww : w $\in$ $\{a, b\}^*$\}. We'll use the pumping lemma. \\ \Large Don't get confused by the use of the variable w both to define L\\ \Large  and as the name for the string we will choose to pump on.\\ \Large  As is always the case, the only real work we have\\ \Large to do is to choose an appropriate string w.\\ \Large We need one that is long enough (i.e., |w| $\geq$ N). \\ \Large And we need one with firm boundaries between regions.\\ \Large So let's choose w = a\^{}Nba\^{}Nb. Since |xy| $\leq$ N, \\ \Large we know that y must occur in the first a region.\\ \Large Clearly if we pump in any additional a's, the two halves of w\\ \Large will no longer be equal. Therefore L is not regular.\end{tabular} \\ \midrule
\begin{tabular}[c]{@{}l@{}}\Large Give a context-free grammar that \\ \Large generate the following language.\\ \Large \{w $\in$ $\{0,1\}^*$, \\ \Large the length of w is odd and the middle symbol is 0\}\\ \\ \textit{\Large Difficulty: Level 2}\\ \textit{\Large Category: Context-Free Languages}\\ \textit{\Large Source: New Jersey Institute of Technology}\end{tabular} & \begin{tabular}[c]{@{}l@{}}\Large G = (V,$\Sigma$,R,S) with set of variables V = \{S\}, where S is the start variable; \\ \Large set of terminals $\Sigma$ = \{0,1\}; and rules S → 0S0 | 0S1 | 1S0 | 1S1 | 0\end{tabular} \\ \midrule
\begin{tabular}[c]{@{}l@{}}\Large Consider the language L = \{$ww^R$\}.\\ \Large Describe a two tape Turing machine to accept L.\\ \\ \textit{\Large Difficulty: Level 3}\\ \textit{\Large Category: Computability Theory}\\ \textit{\Large Source: The University of Texas at Austin}\end{tabular} & \begin{tabular}[c]{@{}l@{}}\Large The two tape machine works as follows: If the input is $\epsilon$, accept.\\ \Large If not, copy the input to the second tape and record in the state\\ \Large that you have processed an even number of characters so far.\\ \Large Now, start the first tape at the left end and the second tape at the right end.\\ \Large Check that the symbols on the two tapes are the same.\\ \Large If not, reject. If so, move the first tape head to the right\\ \Large and the second tape head to the left. \\ \Large Also record that you have processed an odd number and continue,\\ \Large each time using the state to keep track of whether\\ \Large you've seen an even or odd number of characters so far.\\ \Large When you reach the end of the input tape, \\ \Large accept if you've seen an even number of characters.\\ \Large Reject if you've seen an odd number. \\ \Large (The even/odd counter is necessary to make sure \\ \Large that you reject strings such as aba.)\end{tabular} \\ \midrule
\begin{tabular}[c]{@{}l@{}}\Large Let A be the set \{x, y, z\} and B be the set \{x, y\}.\\ \Large What is A × B?\\ \\ \textit{\Large Difficulty: Level 1}\\ \textit{\Large Category: Fundamental Concepts}\\ \textit{\Large Source: Sipser Book}\end{tabular} & \Large A x B = \{(a, b) : a $\in$ A and b $\in$ B\} = \{(x, x), (x, y), (y, x), (y, y), (z, x), (z, y)\} \\ \midrule
\begin{tabular}[c]{@{}l@{}}\Large What is the pumping lemma for regular languages?\\ \\ \textit{\Large Difficulty: Axiomatic}\\ \textit{\Large Category: Theoretical Concepts}\\ \textit{\Large Source: Claude3 + human}\end{tabular} & \begin{tabular}[c]{@{}l@{}} \Large The pumping lemma for regular languages states that if A is a regular language,\\ \Large then there exists a number p such that for any string s in A \\ \Large with length greater than or equal to p, there exist strings x, y, and z where s can\\ \Large be written as xyz, satisfying the following conditions:\\ \Large (1) $xy^iz$ belongs to A for each i greater than or equal to 0,\\ \Large (2) y is not an empty string, and \\ \Large (3) the length of xy is less than or equal to p.\end{tabular} \\ \bottomrule
\end{tabular}
}
\caption{Sample Instances from the TuringQ Dataset}
\label{tab:Instances of TuringQ dataset}
\end{table*}

\begin{table*}[]
\renewcommand{\arraystretch}{1.4}
\resizebox{2\columnwidth}{!}{%
\begin{tabular}{cl}
\toprule
Category & \multicolumn{1}{c}{Description} \\ \midrule
\begin{tabular}[c]{@{}c@{}}Regular\\  Languages\end{tabular} & \begin{tabular}[c]{@{}l@{}}Regular languages are a class of formal languages that can be defined by regular expressions\\ or recognized by finite automata with a finite number of states.\\ Key topics in this category include determining whether a given language is regular or not,\\ often employing techniques like the pumping lemma \\ or constructing regular expressions. Additionally, concepts like deterministic finite automata (DFAs)\\ and nondeterministic finite automata (NFAs) are fundamental, \\ addressing the recognition of strings in regular languages through state transitions based on an input alphabet.\end{tabular} \\ \midrule
\begin{tabular}[c]{@{}c@{}}Context-Free\\  Languages\end{tabular} & \begin{tabular}[c]{@{}l@{}}A context-free language is a formal language that can be precisely defined by a context-free grammar,\\ which consists of a set of production rules specifying how strings of symbols can be derived or generated,\\ regardless of the context in which the symbols appear. Key concepts in the study\\ of context-free languages include context-free grammars themselves,\\ the processes of derivation and parse trees for visualizing derivations,\\ as well as techniques for proving whether a given language is context-free or not.\end{tabular} \\ \midrule
\begin{tabular}[c]{@{}c@{}}Computability\\  Theory\end{tabular} & \begin{tabular}[c]{@{}l@{}}Computability Theory is a branch of theoretical computer science\\ that deals with the limitations and capabilities of computational models,\\ particularly in determining which problems are computationally solvable and which are not.\\ Core concepts include Turing machines, decidability,\\ Turing-recognizable languages, the Church-Turing thesis, and undecidability.\end{tabular} \\ \midrule
\begin{tabular}[c]{@{}c@{}}Complexity\\  Theory\end{tabular} & \begin{tabular}[c]{@{}l@{}}Complexity Theory is a branch of computer science that classifies computational problems\\ based on their inherent difficulty and resource requirements.\\ It analyzes time and space complexity using notations like Big O,\\ and categorizes problems into complexity classes such as P, NP, NP-Complete, and PSPACE.\\ Key concepts include polynomial time solvability, NP-Completeness for hardest problems in NP,\\ and reducibility for relating problem complexities.\end{tabular} \\ \midrule
\begin{tabular}[c]{@{}c@{}}Countability\\  Concepts\end{tabular} & \begin{tabular}[c]{@{}l@{}}Countability concepts revolve around distinguishing between countable and uncountable sets,\\ as well as characterizing the sizes of infinite sets. Key ideas include \\ countable vs. uncountable sets, cardinal numbers and\\ infinite cardinals, bijections and enumeration techniques, diagonalization methods\\ for proving uncountability, the notion of cardinality as a measure of set size,\\ and combinatorial principles like combinations and permutations. \\ These concepts from set theory, combinatorics,\\ and measure theory are crucial for understanding the nature of infinity.\end{tabular} \\ \midrule
\begin{tabular}[c]{@{}c@{}}Fundamental\\  Concepts\end{tabular} & \begin{tabular}[c]{@{}l@{}}Fundamental Concepts are the essential and introductory topics,\\ including Set Theory, Propositional and Predicate Logic, and Relations.\\ Set Theory covers sets, operations, and relations.\\  Logic encompasses logical operators, truth tables, well-formed formulas, and quantifiers.\\ Relations involve properties like reflexivity, symmetry, transitivity, equivalence relations, and partitions.\end{tabular} \\ \midrule
\begin{tabular}[c]{@{}c@{}}Theoretical\\  Concepts\end{tabular} & \begin{tabular}[c]{@{}l@{}}Theoretical Concepts in the theory of computation comprise the principles, theorems,\\ rigorous proofs, lemmas, and auxiliary results that constitute the backbone of the field.\\ These concepts lay the groundwork, illuminate pivotal results through meticulous derivations,\\ and foster a profound understanding by elucidating connections and delineating boundary conditions.\\ Mastering these Theoretical Concepts equips one with a robust theoretical foundation.\end{tabular} \\ \bottomrule
\end{tabular}
}
\caption{Details and Interpretation of the TuringQ Dataset Categories}
\label{tab:categoires}
\end{table*}

\end{document}